\documentclass[conference]{IEEEtran}
\IEEEoverridecommandlockouts
\usepackage{cite}
\usepackage{amsmath,amssymb,amsfonts}
\usepackage{algorithmic}
\usepackage{graphicx}
\usepackage{textcomp}
\usepackage{hyperref}
\usepackage{todonotes}
\usepackage{float}
\usepackage[binary-units=true]{siunitx}
\usepackage{booktabs}
\usepackage[]{subcaption}
\usepackage{standalone}
\usepackage{placeins}
\usepackage{pgfplots}
\usepackage{comment}
\usetikzlibrary{pgfplots.statistics}

\def\BibTeX{{\rm B\kern-.05em{\sc i\kern-.025em b}\kern-.08em
    T\kern-.1667em\lower.7ex\hbox{E}\kern-.125emX}}

\makeatletter
\newcommand{\linebreakand}{%
  \end{@IEEEauthorhalign}
  \hfill\mbox{}\par
  \mbox{}\hfill\begin{@IEEEauthorhalign}
}
\makeatother   
    
\begin{document}

\title{RGB-L: Enhancing Indirect Visual SLAM using LiDAR-based Dense Depth Maps}

\author{\IEEEauthorblockN{Florian Sauerbeck}
\IEEEauthorblockA{\textit{Institute of Automotive Technology } \\
\textit{Technical University of Munich}\\
85748 Garching, Germany  \\
florian.sauerbeck@tum.de}
\and
\IEEEauthorblockN{Benjamin Obermeier}
\IEEEauthorblockA{\textit{Institute of Automotive Technology } \\
\textit{Technical University of Munich}\\
85748 Garching, Germany  \\
benjamin.obermeier@tum.de}
\and
\IEEEauthorblockN{Martin Rudolph}
\IEEEauthorblockA{\textit{Institute of Automotive Technology } \\
\textit{Technical University of Munich}\\
85748 Garching, Germany  \\
martin.rudolph@tum.de}
\linebreakand
\IEEEauthorblockN{Johannes Betz}
\IEEEauthorblockA{\textit{Autonomous Vehicle Systems Lab} \\
\textit{Technical University of Munich}\\
85748 Garching, Germany  \\
johannes.betz@tum.de}\\
}

\IEEEoverridecommandlockouts
\IEEEpubid{\makebox[\columnwidth]{978-1-5386-5541-2/18/\$31.00~\copyright2018 IEEE \hfill} \hspace{\columnsep}\makebox[\columnwidth]{ }}

\maketitle

\IEEEpubidadjcol

\begin{abstract}
In this paper, we present a novel method for integrating 3D LiDAR depth measurements into the existing \textit{ORB-SLAM3} by building upon the RGB-D mode. We propose and compare two methods of depth map generation: conventional computer vision methods, namely an inverse dilation operation, and a supervised deep learning-based approach. We integrate the former directly into the \textit{ORB-SLAM3} framework by adding a so-called RGB-L (LiDAR) mode that directly reads LiDAR point clouds. The proposed methods are evaluated on the KITTI Odometry dataset and compared to each other and the standard \textit{ORB-SLAM3} stereo method. We demonstrate that, depending on the environment, advantages in trajectory accuracy and robustness can be achieved. Furthermore, we demonstrate that the runtime of the \textit{ORB-SLAM3} algorithm can be reduced by more than \SI{40}{\percent} compared to the stereo mode. \\
The related code for the \textit{ORB-SLAM3} RGB-L mode will be available as open-source software under \url{https://github.com/TUMFTM/ORB_SLAM3_RGBL}.
\end{abstract}



\section{Introduction} \label{sec:introduction}
Robust and precise localization is needed for all modules of autonomous vehicle software, such as path planning, trajectory following, or object prediction. To enable full self-driving capabilities, a reliance on satellite-based global localization systems such as the Global Positioning System (GPS) is not feasible. First, such sensors are expensive; second, sensor dropouts can lead to subsequent system failures. Therefore, simultaneous localization and mapping (SLAM) algorithms are important in developing robots and autonomous vehicles \cite{bailey2006, dissanayake2001}. These algorithms can provide a precise robot pose by only using sensor measurements of the environment and therefore provide an important localization technique. To achieve robust SLAM algorithms, it is necessary to include the environmental information from various sensors such as LiDARs and cameras. Since all sensors have their own individual advantages and disadvantages, sensor fusion can be a huge benefit for the localization of autonomous vehicles \cite{marti2019}.

This paper presents an enhancement to the well-known \textit{ORB-SLAM3} \cite{campos2021} algorithm. Currently, this algorithm relies solely on camera data to calculate the robot's pose. The goal is to enhance this algorithm with LiDAR sensor depth measurements to achieve a higher localization accuracy and better robustness e.g. in urban environments. To summarize, this paper comprises four main contributions:

\begin{itemize}
    \item We present a method to integrate LiDAR depth measurements into the existing \textit{ORB-SLAM3} algorithm.
    \item We propose and compare two methods of dense depth map generation from LiDAR point clouds.
    \item We present a variety of experiments for localization of an autonomous vehicle that demonstrates the improved accuracy and robustness of our method.
    \item We compare the runtimes and show a decrease of more than \SI{40}{\percent} compared to stereo mode.
\end{itemize}

\section{Related work} \label{sec:related_work}
Some approaches for fusing multimodal sensor data in SLAM algorithms have been presented in the past and are summarized below.

\subsection{Visual SLAM}
In general, visual SLAM methods can be divided into two subcategories: feature-based (indirect) SLAM and direct SLAM methods. Indirect SLAM tries to detect and match features across multiple image frames and minimize the geometric error. The most-used indirect SLAM is the ORB-SLAM \cite{mur2015, mur2017, campos2021}. \\
In contrast, direct approaches try to directly minimize the photometric error. The most famous direct SLAM algorithms are LSD (Large-Scale Direct Monocular SLAM) \cite{engel2014}, SVO (Semi-direct Visual-Inertial Odometry) \cite{forster2016} and DSO (Direct Sparse Odometry) \cite{engel2017, wang2017}.

\subsection{RGB-D SLAM}
RGB-D (depth) cameras provide an infrared-based depth-measuring approach. Thus, the SLAM algorithm does not have to reconstruct 3D information purely from 2D images. An enhancement of DSO with RGB-D measurements was presented in \cite{yin2022}. The main disadvantage of RGB-D cameras is the limited range of only a few meters and the high dependency on lighting conditions \cite{henry2012}. This makes them unusable for automotive applications.

\subsection{Visual-LiDAR SLAM}
Debeunne et al. \cite{debeunne2020} gave a detailed overview of the state-of-the-art of visual-LiDAR SLAM. Generally, the authors distinguished between concurrent approaches, LiDAR-assisted visual SLAM, camera-assisted LiDAR SLAM, and approaches with tightly coupled sensor fusion. 

\subsubsection{LiDAR Assisted Visual SLAM}
LIMO (Lidar-Monocular) SLAM \cite{graeter2018} uses depth information from LiDAR to fit local planes to obtain depth information for features detected in the camera images. Moreover, semantic information is included.
Shin et al. \cite{shin2018} combined a direct visual SLAM approach with sparse depth data from LiDARs. They used a sliding window-based optimization and included loop closure functionality. Robust SLAM results were obtained on different large-scale datasets. \\
VLOAM (Visual-LiDAR Odometry And Mapping) \cite{zhang2018} uses LiDAR data to get the 3D information of the visual features. Pose estimates from visual feature matching and LiDAR scan matching are combined in graph optimization. The VIL-SLAM, presented by Shao et al. in 2019 \cite{shao2019} combines a visual-inertial odometry (VIO) frontend with LiDAR-mapping and LiDAR-enhanced loop closure. Both algorithms demonstrated state-of-the-art precision under normal conditions with improved robustness against adverse conditions.
The idea to of combining the \textit{ORB-SLAM3} with LiDAR data had previously been presented in \cite{zijlmans2017}, however, no different upsampling methods were presented and a scientific analysis and discussion of the results were missing.

\subsubsection{Tightly Coupled Visual-LiDAR SLAM}
Recently, research has focused on SLAM algorithms that tightly fuse multimodal sensor data. The LIC-fusion method (LiDAR-inertial-camera) \cite{zuo2019} was presented in 2019. It fuses LiDAR features, camera features, and high-frequency IMU data for improved localization accuracy and robustness. Therefore, LIC incorporates online spatiotemporal calibration. Its successor, LIC-fusion 2.0 \cite{zuo2020} added sliding-window plane-feature tracking to further improve performance. \\
Chou et al. \cite{chou2021} presented another tightly coupled visual-LiDAR SLAM: TVL-SLAM. This has separate frontends for LiDAR and camera (ORB-SLAM2-based), which are fused in a shared graph optimization backend with LiDAR loop closure. TVL-SLAM could show improvements over single visual and LiDAR SLAM implementations. \\
R$^{2}$ LIVE \cite{lin2021} uses an error-state Kalman filter in combination with factor-graph optimization to fuse LiDAR, IMU and visual measurements. R$^{3}$ LIVE \cite{lin2022} added a colored 3D map by combining VIO (visual-inertial odometry) and LIO (laser inertial odometry) subsystems. Moreover, online photometric calibration was added to the system. Both algorithms are robust enough to handle various camera failures or LiDAR-degenerated scenarios and can run in real-time on small onboard platforms. \\
Zhao et al. proposed Super Odometry \cite{zhao2021super}, an IMU-centric robust state estimation pipeline. They showed convincing results, especially for perceptually-degraded environments.

\subsection{Generation of Depth Maps}
Our proposed method uses dense depth maps to obtain pixel-wise depth information for the corresponding monocular camera images. Different approaches exist for the generation of such maps. Currently, a major area of research is monocular depth estimation without depth measurements. The current state-of-the-art is presented in \cite{zhao2020, bhoi2019, ming2021}. \\
In this study, we focus on depth completion of sparse depth measurements, namely LiDAR point clouds. To complete sparse depth data with camera images, a distinction is made between two basic approaches: conventional computer vision (CV) and deep learning-based approaches. The approach used by Ku et al. \cite{ku2018} belongs to the former. Their work showed that well-designed classical CV algorithms could outperform deep learning methods under certain circumstances. A significant advantage is that their algorithms require very small resources and can easily run on a CPU. Also, no training is needed, thus overfitting is not possible. \\
By contrast, deep learning-based algorithms dominate in the renowned depth completion challenges such as the KITTI depth completion challenge \cite{Uhrig2017}.
Some of the best-performing and most used networks for LiDAR depth completion are the PENet and ENet \cite{hu2021}, the ACMNet \cite{zhao2021} and the RigNet \cite{yan2021}. Further detailed information on existing approaches and the current state-of-the-art can be found in specific literature review papers \cite{masoumian2022, hu2022}.

\section{Approach}
\begin{figure*}[h!]
		\centering
		\begin{subfigure}[b]{1.0\linewidth}
			\centering
			\includegraphics[width=1.0\linewidth, trim={0cm 0cm 0cm 0cm}, clip]{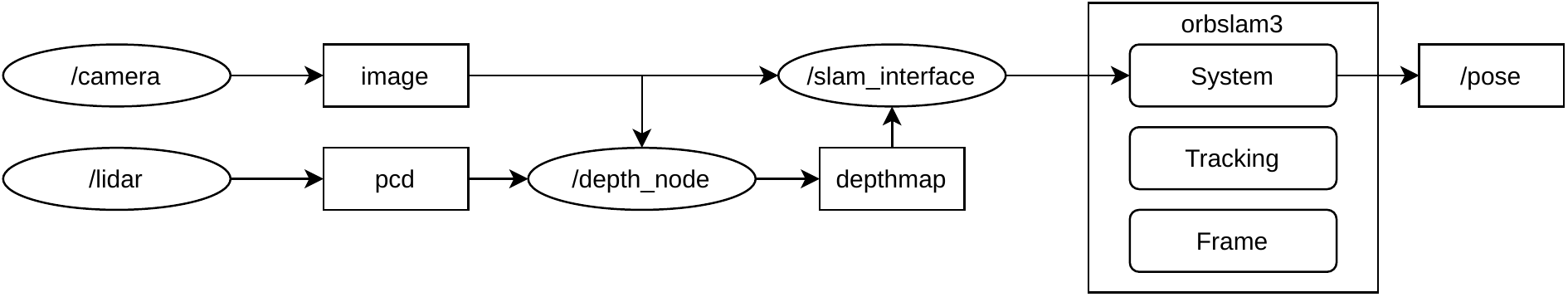}
			\caption{Standard \textit{ORB-SLAM3} RGB-D mode with external depthmap generation}
			\label{fig:rgb-d}
		\vspace{4mm}
		\end{subfigure}
		\begin{subfigure}[b]{1.0\linewidth}
			\centering
			\includegraphics[width=1.0\linewidth, trim={0cm 0cm 0cm 0cm}, clip]{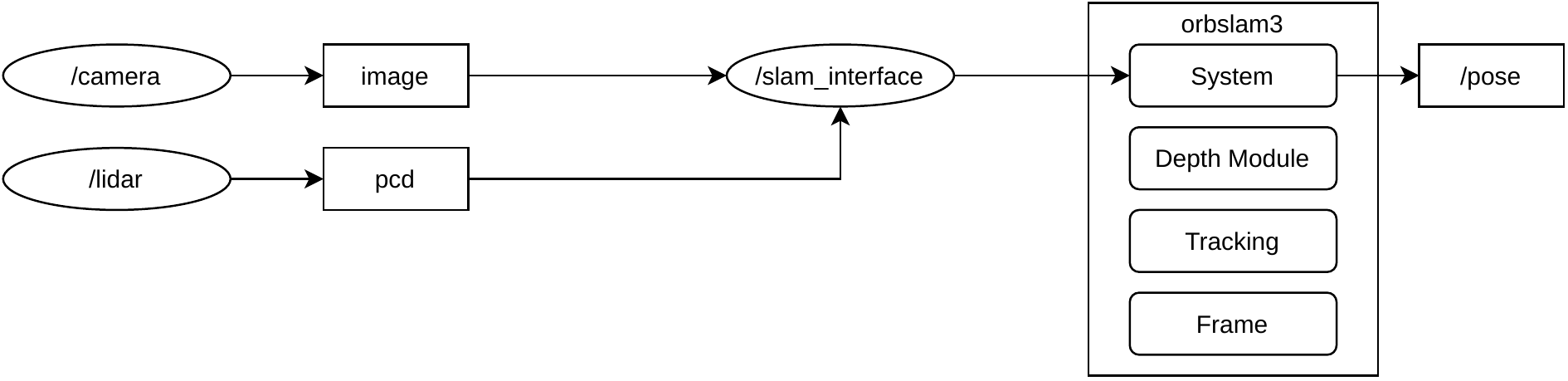}
			\caption{Novel RGB-L mode with integrated depth module}
			\label{fig:rgb-l}
		\end{subfigure}	
		\caption{Setup of the ORB-SLAM3 for working with LiDAR-based dense depth maps in the standard RGB-D mode and the novel RGB-L mode.}
		\label{fig:overview}
	\end{figure*}

The overall approach of this work is presented in Figure \ref{fig:overview}. The camera frame is handed over to the SLAM interface. LiDAR depth measurements (point cloud data (pcd)) are transformed into the camera frame and upsampled.

\subsection{ORB SLAM Integration}
As a baseline indirect visual SLAM algorithm, we use ORB-SLAM3 \cite{campos2021}. An RGB-D mode is already integrated and provides the option of directly feeding depth maps to each corresponding camera image. This allows us to use the RGB-D mode with previously generated and saved depth maps as depicted in Figure \ref{fig:rgb-d}.
Moreover, we introduce RGB-L mode, which upsamples LiDAR data online using conventional CV and feeds the results to the SLAM module. The computation graph is shown in Figure \ref{fig:rgb-l}. The \textit{depth\_node} transforms LiDAR data into the camera frame and upsamples the data online using conventional CV methods.

\subsection{Generation of Depth Maps}
\begin{figure}[H]
	\centering
		\includegraphics[width=0.9\linewidth, trim={0cm 0cm 0cm 0cm}, clip]{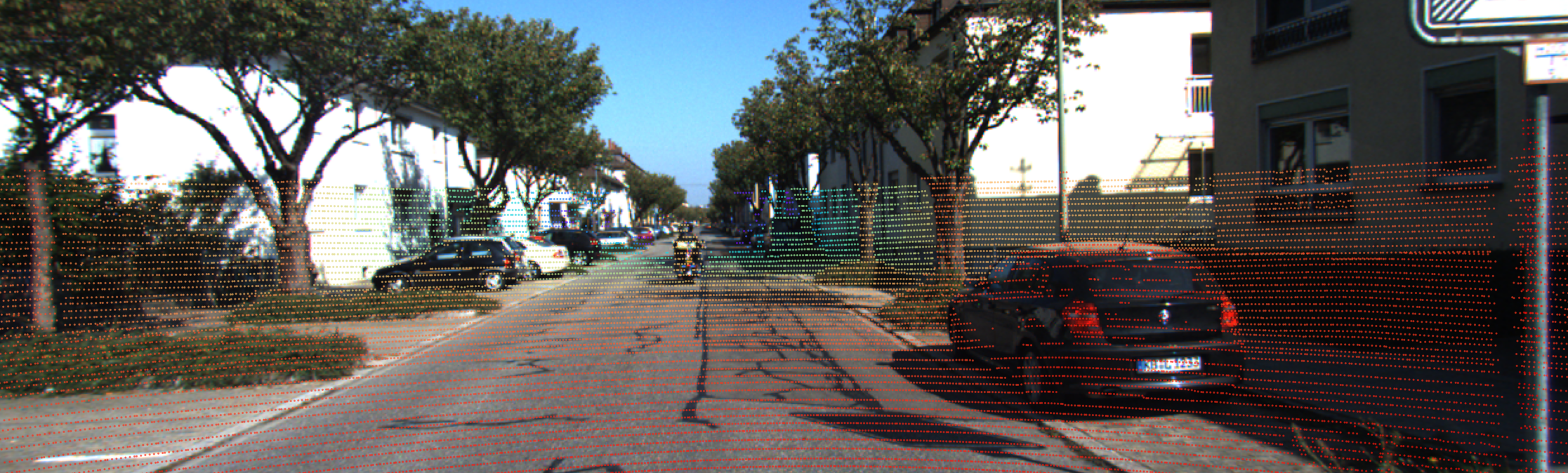}
		\caption{Projection of sparse LiDAR point cloud into camera image.}
		\label{fig:projection}
\end{figure}

When compared to cameras, LiDARs output sparse data. Thus, it is difficult to get pixel-level depth information which is needed to transform visual features into 3D space. Figure \ref{fig:projection} shows a LiDAR point cloud transformed into the corresponding camera image. \\

We propose two different methods of depth map generation, which will be evaluated and compared below: a conventional CV approach and a deep learning-based method. Multiple experiments with different conventional upsampling methods have revealed that an inverse dilation with a 5x5 diamond kernel exhibits the best results amongst conventional CV methods. Usually, image dilation is used to increase the size of the thickness of foreground objects. Here, we want to use it to increase the density of the LiDAR point cloud transformed into the image frame. The kernel matrix we chose is of size 5x5 pixels and has a diamond shape. \\
The unprocessed depth maps from the KITTI Odometry dataset have an average sparsity of around \SI{96}{\percent} after projection into the camera frame. By applying the inverse dilation, we can decrease the sparsity to around \SI{65}{\percent} The generated depth map has an MAE of \SI{1.03}{\meter} and an RMSE of \SI{4.41}{\meter}. We found that methods with denser outputs tend to have worse tracking results because the depth errors increase in between the single depth measurements. \\

\begin{figure}[h]
		\includegraphics[width=0.85\linewidth, trim={0cm 0cm 1.4cm 0cm}, clip]{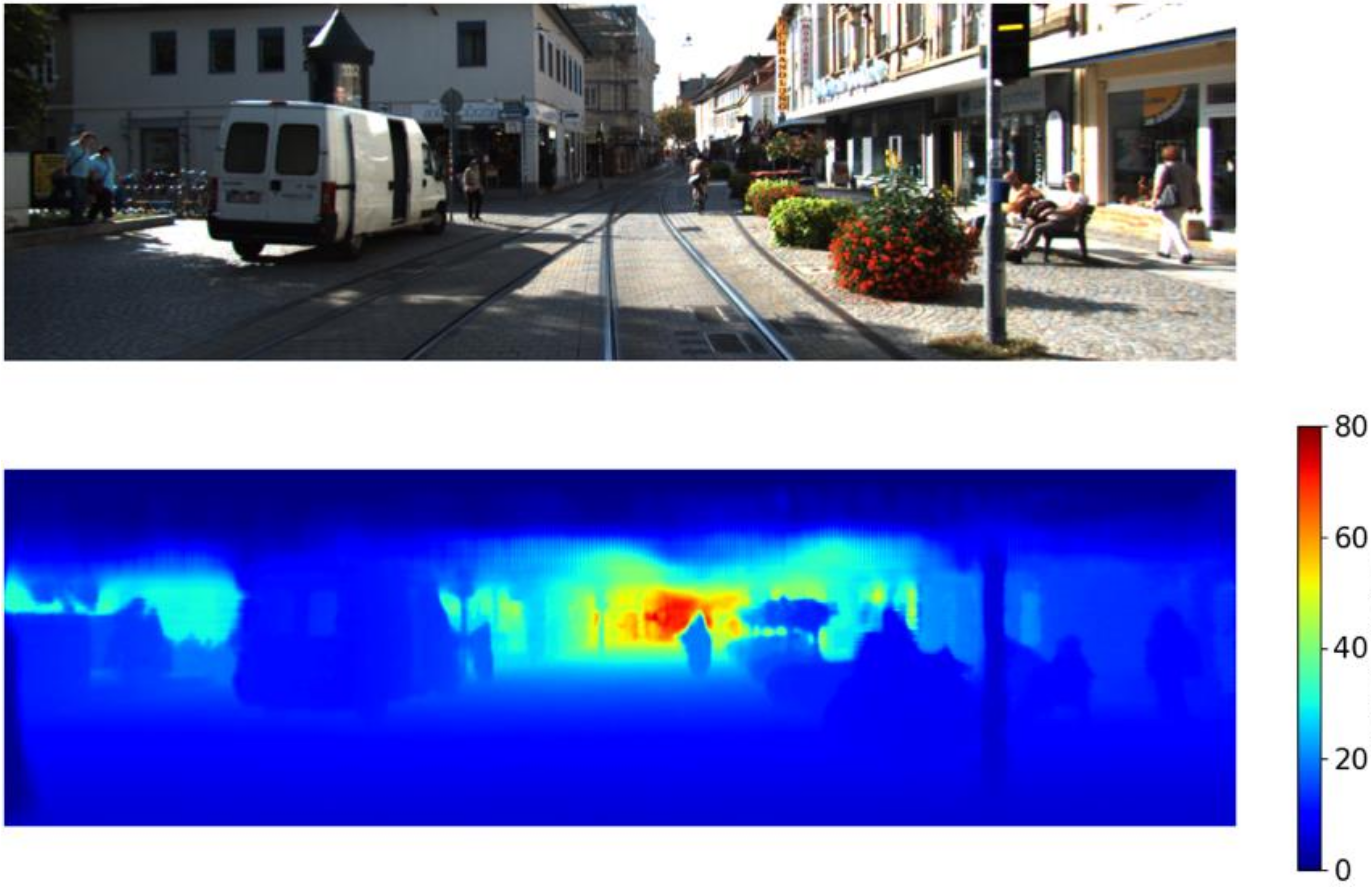}
		\put(16,16){\rotatebox{90}{Depth in \si{\metre}}}
		\put(2,70){80}
		\put(2,53){60}
		\put(2,35){40}
		\put(2,17){20}
		\put(2,0){0}
		\caption{Typical camera image and corresponding depth map generated by the ENet based neural network.}
		\label{fig:depth_map}
\end{figure}
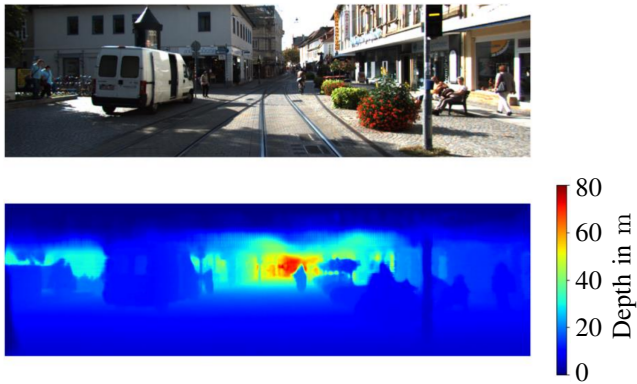

With deep learning methods, we can achieve a density of \SI{100}{\percent}. This means that each pixel has a corresponding depth value. However, those values are not worthwile for the area without LiDAR returns because the KITTI depth dataset does not contain ground truth for those pixels. For the deep learning-based LiDAR depth completion, we use a CNN based on the ENet \cite{hu2021} which we have previously optimized for runtime. In our configuration, the network has an average runtime of \SI{16.32}{\milli\second}, \SI{9.66}{\milli\second} of which are the actual computation time on the GPU. For the MAE per pixel we achieve values of \SI{0.39}{\meter} and for the RMSE \SI{1.17}{\meter}. Those values were generated on the KITTI Depth dataset. Values for the Odometry data might differ. However, since the data comes from the same source, it is a justified assumption that the accuracy of the Odometry data is comparable.
A typical high resolution depth map from the neural network and the corresponding camera image are shown in Figure \ref{fig:depth_map}. \\

The depth in the image area for which we did not have ground truth data for training the network was not used. However, it is hard to find a general rule about which areas to use and which to ignore.

\section{Results}
This section summarizes the results we achieved with the presented method. We want to compare our results to the \textit{ORB-SLAM3} stereo module in terms of accuracy, and particularly in terms of the runtime. The experiments were conducted on the KITTI Odometry dataset \cite{geiger2013}.

\subsection{Comparison of the Trajectories}
Tables \ref{tab:Results_KITTI_translation} and \ref{tab:Results_KITTI_rotation} show the translational and rotational error of the mentioned SLAM modes on the ten KITTI Odometry sequences. \\
Stereo and RGB-L exhibit comparable performances with a slight advantage for the stereo mode. It can be seen directly that the accuracy of the deep learning methods is significantly worse for both, translational and rotational errors. This might be because it is hard to generally define areas, where the net performs well. To minimize the negative effect of this, the upper \SI{30}{\percent} of the depth maps were ignored. Another downside of the neural net is that the KITTI ground truth only uses values of up to \SI{80}{\meter}. Thus, even areas that are significantly further away will show depth values around \SI{80}{\meter}. The RGB-L mode only uses depth values that are close to the actual depth measurements from the LiDAR and thus generalizes better.

\begin{figure}[tb!]
	\centering
	\begin{subfigure}[b]{0.9\linewidth}
		\centering
		\includegraphics[width=\linewidth, trim={0cm 0cm 0cm 0cm},clip]{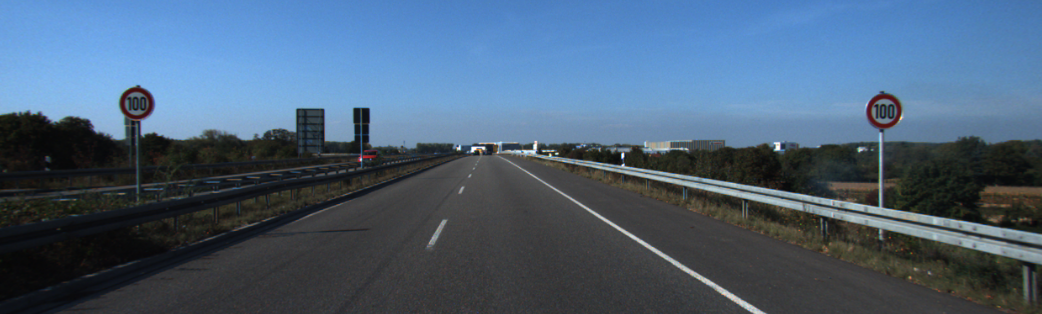}
		\caption{KITTI Sequence 01, Frame 248}
	\end{subfigure}
	\par \smallskip
	\begin{subfigure}[b]{0.9\linewidth}
		\centering
		\includegraphics[width=\linewidth, trim={0cm 0cm 0cm 0cm},clip]{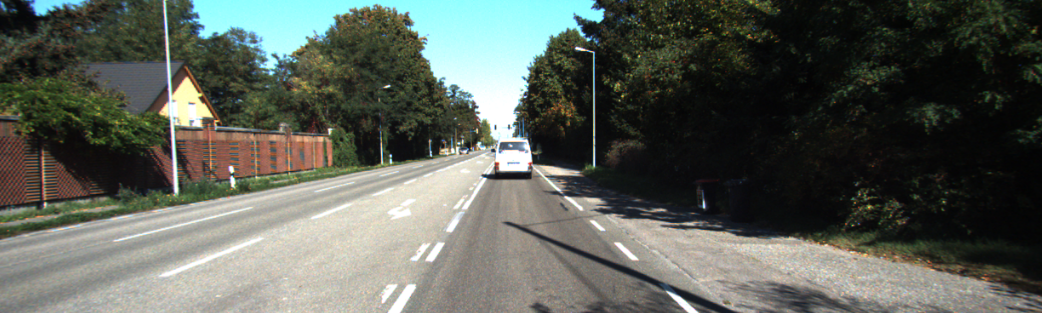}
		\caption{KITTI Sequence 04, Frame 152}
	\end{subfigure}
	\caption{Typical images of low-feature environments. RGB-L beats ORB3 stereo mode accuracy here.}
	\label{fig:low-feature}
\end{figure}

\begin{table}
\caption{Average translational error in \SI{}{\percent}. Optimum results are highlighted in green and the next best in blue. \label{tab:Results_KITTI_translation}}
\begin{center}
	\begin{tabular}{ll|ccc}
		\toprule
		Seq. & Scenery  & Stereo & RGB-L & Deep Learning  \\
			\midrule
			& & \multicolumn{3}{c}{Translational Error in \%}\\
			\midrule
			00 & Urban 			& \textcolor{blue}{0.704}	& \textcolor{olive}{0.695} 	& 0.721  \\
			01 & Highway 		& \textcolor{blue}{1.628}	& \textcolor{olive}{1.098} & 1.698  \\
			02 & Urban 			& \textcolor{olive}{0.817} & \textcolor{blue}{0.836} 	& 0.852  \\
			03 & Urban 			& \textcolor{blue}{0.950} 	& 1.033 					& \textcolor{olive}{0.726}  \\
			04 & Rural 			& \textcolor{blue}{0.562}	& \textcolor{olive}{0.453} 	& 1.121  \\
			05 & Urban 			& \textcolor{olive}{0.413} & \textcolor{blue}{0.470} 	& 0.840  \\
			06 & Urban 			& \textcolor{olive}{0.544} & \textcolor{blue}{0.878}	& 1.182  \\
			07 & Urban 			& \textcolor{olive}{0.483} & \textcolor{blue}{0.581} 	& 0.862  \\
			08 & Urban 			& \textcolor{olive}{0.991} & \textcolor{blue}{1.085} 	& 1.284  \\
			09 & Rural 			& \textcolor{blue}{0.988}	& \textcolor{olive}{0.866} 	& 1.092  \\
			10 & Urban 			& \textcolor{olive}{0.710} & \textcolor{blue}{0.840} 	& 1.388  \\
			\midrule
			\multicolumn{2}{c}{Average}& 0.800 & 0.801 & 1.070  \\	
		\bottomrule
	\end{tabular}
\end{center}
\end{table}

\begin{table}
\caption{Average rotational error in \SI{}{\degree}/\SI{100}{\meter}. Optimum results are highlighted in green and the next best in blue. \label{tab:Results_KITTI_rotation}}
\begin{center}
	\begin{tabular}{ll|ccc}
		\toprule
		Seq. & Scenery  & Stereo & RGB-L & Deep Learning \\
		& & \multicolumn{3}{c}{Rotational Error in °/100m} \\
			\midrule
			00 & Urban 			& \textcolor{blue}{0.272} & \textcolor{olive}{0.257} & 0.315 \\
			01 & Highway 		& \textcolor{olive}{0.201} & \textcolor{blue}{0.364} & 0.681 \\
			02 & (Sub-) Urban 	& \textcolor{blue}{0.275} & \textcolor{olive}{0.259} & 0.344 \\
			03 & Urban 			& \textcolor{olive}{0.235} & \textcolor{blue}{0.258} & 0.334 \\
			04 & Rural 			& \textcolor{blue}{0.251} & \textcolor{olive}{0.221} & 0.985 \\
			05 & Urban 			& \textcolor{olive}{0.158} & \textcolor{blue}{0.214} & 0.419 \\
			06 & Urban 			& \textcolor{olive}{0.224} & \textcolor{blue}{0.466} & 0.425 \\
			07 & Urban 			& \textcolor{olive}{0.271} & \textcolor{blue}{0.293} & 0.497 \\
			08 & Urban 			& \textcolor{olive}{0.303} & \textcolor{blue}{0.338} & 0.556 \\
			09 & Rural 			& \textcolor{olive}{0.299} & \textcolor{blue}{0.342} & 0.438 \\
			10 & Urban 			& \textcolor{olive}{0.343} & \textcolor{blue}{0.421} & 0.484 \\
			\midrule
			\multicolumn{2}{c}{Average}& 0.244 & 0.312 & 0.498  \\	
		\bottomrule	
	\end{tabular}
\end{center}
\end{table}

Upon further investigation of the accuracy of the algorithms across the different sequences, it becomes apparent that the RGB-L mode performs especially well in low-feature environments, as shown in Figure \ref{fig:low-feature}.

\subsection{Runtime Evaluation}
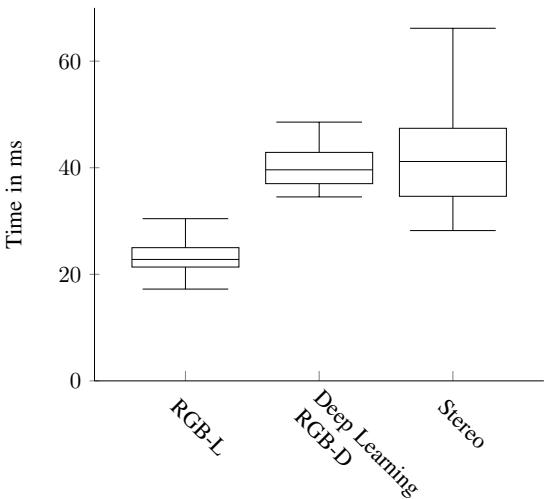
\begin{figure}[!htbp]
\centering
	\definecolor{tumblue}{rgb}{0,0.39,0.74}
\definecolor{tumgreen}{rgb}{0.64, 0.68, 0}
\definecolor{tumorange}{rgb}{0.89, 0.45, 0.13}
\definecolor{tumgray}{rgb}{0.6, 0.6, 0.6}

\begin{tikzpicture}
\begin{axis}[
boxplot/draw direction=y,
axis x line*=bottom,
axis y line=left,
y axis line style=-,
x axis line style=-,
xtick={1,2,3},
xticklabels={RGB-L, Deep Learning \\ RGB-D, Stereo},
x tick label style={
	rotate=-45,anchor=west,
	yshift=-10pt,
	text width=5cm},
ymin=0, ymax=70,
ylabel=Time in ms
]
\addplot+ [
color=black,
boxplot prepared={
	lower whisker=17.226, 
	lower quartile=21.362,
	median=22.79,
	upper quartile=24.998, 
	upper whisker=30.438,
},
] coordinates {};

\addplot+ [
color=black,
boxplot prepared={
	lower whisker=34.524,
	lower quartile=37.013,
	median=39.608,
	upper quartile=42.879, 
	upper whisker=48.562,
},
] coordinates {};

\addplot+ [
color=black,
boxplot prepared={
	lower whisker=28.205, 
	lower quartile=34.637,
	median=41.17,
	upper quartile=47.402, 
	upper whisker=66.173,
},
] coordinates {};

\end{axis}
\end{tikzpicture}
	\vspace*{-25mm}
	\caption{Runtime comparison of the presented methods.}
	\label{fig:runtime}
\end{figure}

As SLAM is a highly time-critical process that has to run in real-time on autonomous vehicles, we want to put a special focus on runtime analysis. \\
Figure \ref{fig:runtime} compares the total runtimes of the three modes. Both modes that use depth maps from LiDAR, show significantly smaller variances. The mean value of the RGB-L runtime is \SI{22.8}{\milli\second} and thus is more than \SI{40}{\percent} less than the \SI{41.2}{\milli\second} mean from stereo mode. \\
The composition of the runtimes is summarized in Table \ref{tab:runtime}. Since the RGB-D mode requires previously generated depth maps, it should be noted that the partial computation times from offline and online calculations were added together. The conventional CV algorithm has a very low computation time and can therefore significantly accelerate the ORB-SLAM3 algorithm in comparison to the stereo mode.

\begin{table*}
\begin{center}
\caption{Response times for stereo mode and the proposed RGB-L mode.}
\label{tab:runtime}
	\begin{tabular}{l|ccc|ccc|ccc}
			\toprule
			Operation & \multicolumn{3}{c|}{RGB-L with 5x5 Dilation} & \multicolumn{3}{c|}{Deep Learning-based} & \multicolumn{3}{c}{Stereo} \\
			& min & median & max & min & median & max & min & median & max\\
			\midrule
			Neural Net GPU		& - & - & - & 9.25 & 9.66 & 10.21 & - & - & - \\
			Neural Net total	& - & - & - & 15.38 & 16.32 & 18.63 & - & - & - \\
			Projection 			& 1.52 & 2.24 & 3.61 & - & - & - & - & - & - \\
			Upsampling 			& 0.43 & 0.57 & 0.83 & - & - & - & - & - & - \\
			Tracking 			& 17.23 & 22.79 & 30.44 & 19.14 & 23.29 & 29.93 & 28.21 & 41.17 & 66.18 \\
			\midrule
			Overall & 19.18 & 25.60 & 34.88 & 34.52 & 39.61 & 48.56 & 28.21 & 41.17 & 66.17 \\
			\bottomrule			
	\end{tabular}
\end{center}
\end{table*}
	
The referred to smaller mean computation time in combination with the smaller variance could make the presented approach particularly interesting for embedded applications with limited hardware resources. The whole of the algorithms runs on a CPU without the need for a power-intensive GPU.

\section{Discussion and Future Work}
The presented algorithms were evaluated using the KITTI Odometry dataset. To generate further results and conclusions, an evaluation using other datasets could well be worthwhile. Since the RGB-L mode demonstrated high accuracy in low-feature environments, also bad weather scenarios or partially polluted camera images could be of special interest.
Since the RGB-L mode does not use deep learning methods, good scalability to other data can be assumed. The chosen algorithms do not depend on the exact camera or LiDAR model, the environment, or the data-base. Also, no expensive ground truth depth data is needed. Only the camera's internal calibration and the external calibration of the camera and LiDAR need to be adjusted. \\
Further research could also be conducted in the field of deep learning-based depth map generation. For example, including a confidence map to decide which depth values best to use in the SLAM algorithm could be highly advantageous. Another problem of the deep learning-based approach is scalability. The depth net will have to be retrained for every new dataset to obtain the best results. This is currently one of the most significant issues that deep learning approaches are facing and will undoubtedly be one of the major research topics in the coming years. \\
Including the deep-learning pipeline with the GPU directly in the RGB-L mode can help to reduce computation time thus making online use possible. \\
The proposed method builds upon existing open-source ORB-SLAM3. We present a simple open-source method that can easily be used by other researchers. However, sensor fusion is not ideal. The ORB-SLAM uses the depth values from the depth maps to generate "quasi-stereo" images for handling the depth information. As already pointed out in section \ref{sec:related_work}, there are some other ways of combining LiDAR and camera data for localization and mapping. Future research on the optimal way to fuse that information will need to be conducted.

\section{Conclusion}
In this paper, we have proposed a method of fusing LiDAR depth measurements into an existing indirect visual SLAM algorithm, namely ORB-SLAM3 \cite{campos2021}. We compare two methods to generate pixel-wise depth maps for each camera image. For upsampling the LiDAR point cloud, we used an inverse dilation with a 5x5 pixel-sized diamond kernel and a supervised deep learning-based method based on ENet \cite{hu2021}. While the deep learning-based depth maps had to be created offline before running the SLAM, the dilation was directly included in the ORB SLAM3 as RGB-L mode. The code will be available as open-source software. \\
Depending on the sequence of the KITTI Odometry dataset, RGB-L mode could match the performance of the stereo method and even outperform it in low-feature environments. The use of deep learning-based depth maps did not prove beneficial in this work and will be the subject of future research. With RGB-L, we were able to reduce the runtime by more than \SI{40}{\percent} when compared to the stereo mode, while at the same time the runtime variance decreased significantly. This makes our approach particularly interesting for low-power and low-latency applications.

\section*{Acknowledgment}
As the primary author, Florian Sauerbeck proposed the initial idea for this paper, and he is responsible for the concept presented and the implementation thereof. Benjamin Obermeier contributed to the generation of the results and the development of the workflow. Martin Rudolph contributed to the development of the conventional LiDAR data upsampling and the integration of the RGB-L mode into \textit{ORB-SLAM3}. Johannes Betz contributed to the conception of the research project and revised the paper critically for important intellectual content. He gave final approval of the version to be published and agreed to all aspects of the work. We thank the ADAC Stiftung and TUM for funding the underlying research project.
We want to thank Marius Spitzar whose master’s thesis contributed to the implemetation of the depth completion neural net.
\FloatBarrier

\bibliographystyle{ieeetr}
\bibliography{./refs}

\end{document}